\pdfoutput=1

\documentclass[11pt]{article}

\usepackage[]{acl}

\usepackage{times}
\usepackage{latexsym}
\usepackage{hyperref}
\usepackage[T1]{fontenc}
\usepackage{csquotes}
\usepackage{amssymb}

\usepackage[utf8]{inputenc}

\usepackage{microtype}
\usepackage{xspace}
\usepackage{amsmath} 
\usepackage{graphicx}

\usepackage{subcaption}
\usepackage{caption}

\usepackage{fixltx2e}

\title{Hitting \enquote{Probe}rty with Non-Linearity, and More}

\author{Avik Pal \\
  University of Amsterdam \\
  \texttt{avik.pal@student.uva.nl} \\\And
  Madhura Pawar \\
  University of Amsterdam \\
  \texttt{madhura.pawar@student.uva.nl} \\}

\begin{document}
\maketitle
\begin{abstract}
Structural probes \cite{Hewitt2019ASP} learn a linear transformation to find how dependency trees are embedded in the hidden states of language models. This simple design may not allow for full exploitation of the structure of the encoded information. Hence, to investigate the structure of the encoded information to its full extent, we incorporate non-linear structural probes. We reformulate the design of non-linear structural probes introduced by \cite{WhitePSC21} making its design simpler yet effective. We also design a visualization framework that lets us qualitatively assess how strongly two words in a sentence are connected in the predicted dependency tree. We use this technique to understand which non-linear probe variant is good at encoding syntactical information. Additionally, we also use it to qualitatively investigate the structure of dependency trees that BERT encodes in each of its layers. We find that the radial basis function (RBF) is an effective non-linear probe for the BERT model than the linear probe.
\end{abstract}

\section{Introduction}



In human languages, the meaning of a sentence is composed hierarchically - small chunks of words together make successively larger chunks. We refer to this tree-structured hierarchy  of a sentence as a dependency tree. 
The paper, \cite{Hewitt2019ASP} introduces \enquote{Structural Probes} that tests a simple hypothesis for how dependency trees may be embedded in the hidden states of a language model. The probe in this context is a model that learns a linear transformation such that two words that are syntactically close to one another should also have less distance between their respective contextual representations.

One of the advantages of these structural probes is simplicity. However, this simple design may not allow for full exploitation of the structure of the encoded information. This motivates us to investigate whether the dependency structure is encoded in a non-linear way.  
\cite{WhitePSC21} introduces three non-linear structural probes - polynomial, radial basis function, and sigmoid. We reformulate these non-linear probe variants. We then apply these probes on contextualized embeddings of BERT and BERT$_\text{LARGE}$ \cite{devlin-etal-2019-bert}. Since different layers of BERT capture different linguistic properties \cite{Rogers2020API}, can our probing method tell us how dependency information is encoded in every layer? This question (as far as we know) hasn't been answered yet in the existing literature.  We answer this question by applying a non-linear probing variant-RBF across the layers of BERT and BERT$_\text{LARGE}$.

We expect the non-linear variants to perform better than linear probes due to the complex nature of the dependency trees which might be better captured by non-linear probes. We test this hypothesis quantitatively using undirected attachment score (UUAS) \cite{Hewitt2019ASP}. Additionally, we test this hypothesis qualitatively by introducing a measure that computes the strength with which the probe predicts dependency between two words. 

We find that Radial Basis Function (RBF) probe is more effective than the linear probe for BERT. And we are also able to understand how every layer gradually encodes syntactical information to form a complete dependency tree. 
\section{Background}
Previous works by \cite{gulordava-etal-2018-colorless, kuncoro-etal-2018-lstms, LinzenL18, futrell-etal-2019-neural} show that Recurrent Neural Networks (RNNs) were able to learn the syntactic structure of languages within their hidden states without being explicitly trained on them. But these experiments could only show local linguistic phenomena owing to the limited potential of the representations formed by looking at the immediate hidden state in an iterative manner that loses distant past information. With the advent of the attention mechanism \cite{BahdanauCB14}, the encoders could form better representations by looking at hidden states of all previous steps enabling the model to capture distant information well. This expedited more robust frameworks to quantitatively align with underlying linguistic structures in hidden states.

Structural probing is one such framework that attempts to provide evidence of syntax trees embedded in hidden states using UUAS \cite{Hewitt2019ASP} while also providing experimental setups to visualize syntactic information learning capability between layers. In the paper \cite{WhitePSC21}, the authors further kernelize this framework to incorporate non-linear formulations for increased expressivity of the hidden states.

Our work is majorly a reformulation of these kernelized non-linear probes with simpler design and effectiveness. 

\section{Approach} \label{sec:app}
\subsection{Structural Probe}
The goal of the structural probe is to see whether the syntactic distance between any two words can be approximated by a learned, linear distance function:
\begin{align} \label{eqn:1}
d_B(h_i^l,h_j^l) = || Bh_i^l - Bh_j^l||_2
\end{align} 
where $h_i^l$ and $h_j^l$ are word embeddings from the $l^{th}$ layer of the language model for $i^{th}$ and $j^{th}$ word where $h_i^l, h_j^l \in \mathbb{R}^{d}$ and $B \in \mathbb{R}^{d \times m}$ is a linear projection matrix. To learn this probe, \cite{Hewitt2019ASP} minimize the following objective with respect to $B$ through gradient descent:
\begin{align} \label{eqn:2}
\min_B \sum_l \frac{1}{|s^l|^2}\sum_{i,j}|d_T^l(h_i^l,h_j^l)-d_B^l(h_i^l,h_j^l)^2|
\end{align} 
where $|s^l|$ is the length of the sentence and $d_T^l(h_i^l,h_j^l)$ is the actual distance between words $i$ and $j$ in the dependency tree of sentence $s$. This minimizes the difference between the syntactic distances obtained from the dependency tree and the distance between the two vectors under our learned transformation. From now on we refer to this linear design of structural probes as \enquote{linear probes}.

\subsection{Non-Linear Probes}
\cite{WhitePSC21} finds the distance between two embeddings as follows: 
\begin{align} \label{eqn:3}
 ||\phi(Bh_i^l,Bh_i^l) - 2\phi(Bh_i^l,Bh_j^l) + \phi(Bh_j^l,Bh_j^l)||_2
\end{align} 
where $\phi$ is the non-linear function that takes as input the linear transformation of two representations. We modify the kernel to make it a function that takes as input just one linear transformation of the representation and computes distance as follows:
\begin{align} \label{eqn:4}
d_B(h_i^l,h_j^l) = ||\phi(Bh_i^l) - \phi(Bh_j^l)||_2
\end{align} 
We design our probe in such a way so that it has a simple design while factoring in the effect of applying non-linearity on the embeddings. 
Firstly we use the polynomial function as follows:
\begin{align} \label{eqn:5}
\phi_{\text{poly}}(Bh_i^l) = (Bh_i^l+c)^d
\end{align} 
 where $d \in \mathbb{Z}_+$ and $c \in \mathbb{R}_{\ge}0$. Next, we consider the radial-basis function (RBF), defined as follows: 
 \begin{align} \label{eqn:6}
\phi_{\text{rbf}}(Bh_i^l) = \exp(-\frac{||Bh_i^l||^2}{2\sigma^2})
\end{align}
 where $\sigma$ acts like a scaling factor. Lastly, we implement the sigmoid function:
 \begin{align} \label{eqn:7}
\phi_{\text{rbf}}(Bh_i^l) = \tanh(aBh_i^l+b)
\end{align}
where $a$ and $b$ are scalar-valued tuning parameters. 
 
\subsection{Tree Distance Evaluation Metric} 
\label{ssec:metric}
 As our quantitative metric, We use Unlabeled Undirected Attachment Score (UUAS) which is the percent of edges predicted correctly by the probe against the actual dependency tree (gold tree). UUAS score merely conveys whether there exists an edge or not between two word embeddings. It would be helpful to also know how strongly the probe thinks two words are connected in a sentence's predicted dependency tree. Inspired by the qualitative visualization introduced by \cite{Coenen2019VisualizingAM}, we implement a similar visualization framework to see how strongly the probe predicts connectivity between $h_i^l$ and $h_j^l$:
 \begin{align} \label{eqn:8}
 \text{strength}(h_i^l,h_j^l) = \frac{d_B(h_i^l,h_j^l)}{d_T(h_i^l,h_j^l)}
\end{align}

This metric can help us visualize qualitatively how different probes encode the dependency trees from the contextual representations. Also, this measure can help us gain insight into how each layer encodes  dependency tree syntaxes. 

\begin{figure}[t]
    \centering
    \includegraphics[width=0.5\textwidth]{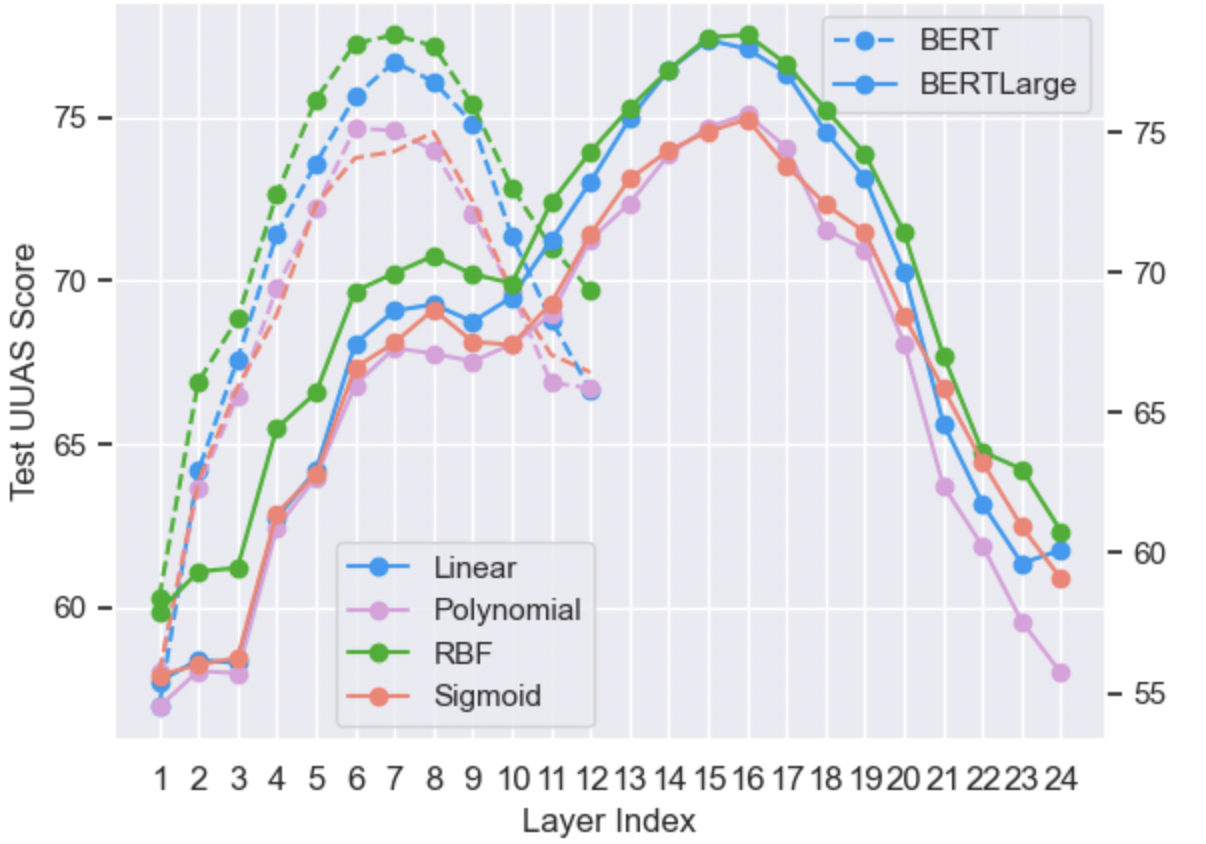}
    \caption{UUAS across BERT and BERT$_\text{LARGE}$  layers}
    \label{fig:bbl}
\end{figure}

\section{Experiments}

\subsection{Dataset}
We use the Universal Dependencies English Web Treebank (UD-EWT) dataset \cite{silveira14gold}. The EWT trees were hand-corrected to Universal Dependencies which have a CoNLL-U format \cite{buchholz-marsi-2006-conll}. We keep the original train-dev-test split of 75.5-12-12.5 provided by \cite{silveira14gold}.

\subsection{Models and Setup}
We use our probes to analyze the dependency structure present in the hidden states of  BERT and BERT$_\text{LARGE}$. We take a contextualized representation of the words in the sentence while taking the mean over all sub-word representations. We keep hyperparameters for Equation \ref{eqn:5}: $c=0, d=2$, Equation \ref{eqn:6}: $\sigma=1$, and Equation \ref{eqn:7}: $a=1, b=0$. We train our probes for 200 epochs with early stopping enabled and an initial learning rate of 0.001 with a reduction factor of 0.5 on plateauing.
The code implementation of our project is here\footnote{\url{https://github.com/madhurapawaruva/nlp2-probing-lms}}.


\begin{figure}
    \centering
    \includegraphics[width=0.5\textwidth]{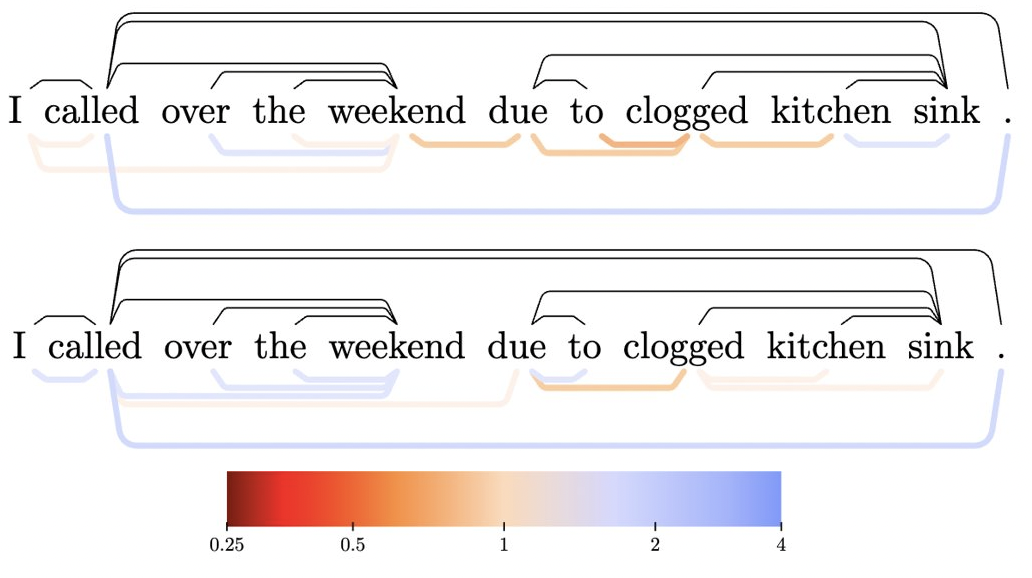}
    \caption{Dependency trees for BERT Layer 12 by Linear Probe (above) and RBF Probe (below). Edges in black are the gold trees.}
    \label{fig:sent1}
\end{figure}

\begin{figure*}

  \centering
  \begin{subfigure}[l]{0.9\textwidth}
    \centering
    \includegraphics[width=\textwidth]{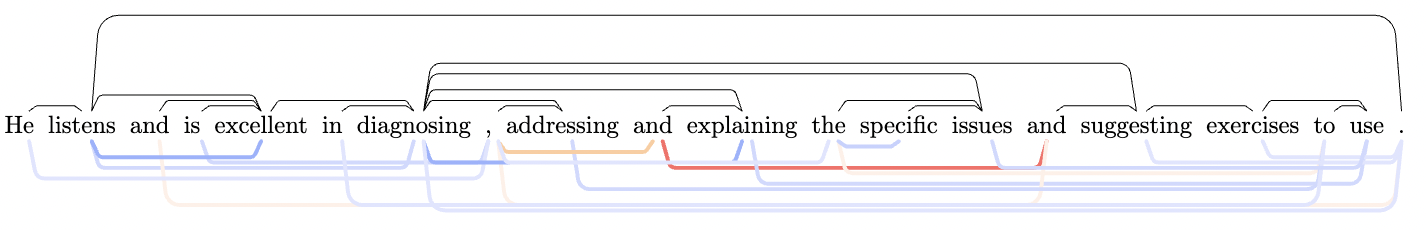}
    \caption{Layer 3 (UUAS = 68.842)}
     \label{fig:a}
  \end{subfigure}
  
  \centering
  \begin{subfigure}[l]{0.9\textwidth}
    \centering
    \includegraphics[width=\textwidth]{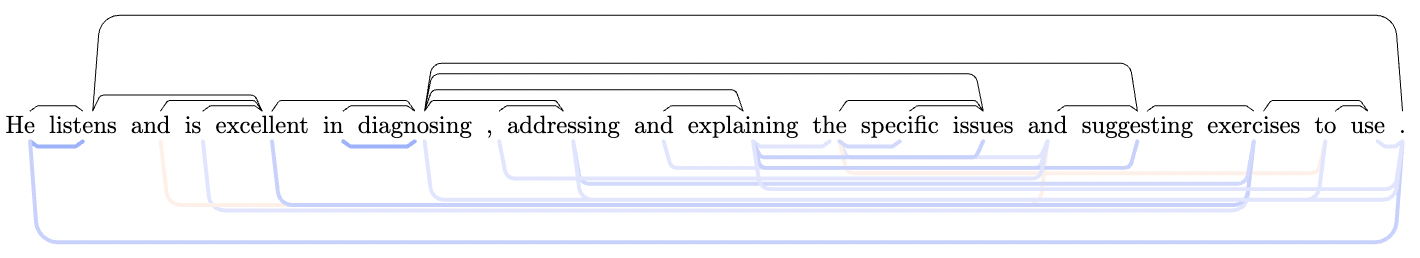}
    \caption{Layer 6 (UUAS = 77.269)}
    \label{fig:b}
  \end{subfigure} 
  
  \centering
  \begin{subfigure}[l]{0.9\textwidth}
    \centering
    \includegraphics[width=\textwidth]{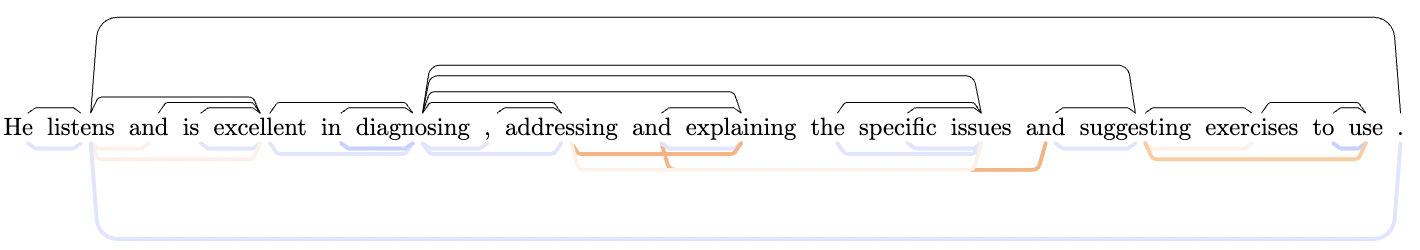}
    \caption{Layer 12 (UUAS = 69.71)}
    \label{fig:c}
  \end{subfigure} 
  ~
 
  \caption{Dependency trees from predicted squared distances on BERT by RBF Probe for layers $\in \{3,6,12\}$}
\end{figure*}

\section{Results and Discussion} \label{sec:discuss}
We initially implemented the equations for non-linear probes introduced by \cite{WhitePSC21}. However, the design of their probes wasn't clear as they don't mention how they find the distance between two embeddings. Because of this, we weren't able to reproduce their work. Hence, based on our intuition and understanding of structural probes we came up with Equations \ref{eqn:4}-\ref{eqn:7}. 

We observe in Figure \ref{fig:bbl} that for BERT and BERT$_\text{LARGE}$, \textbf{RBF Probe has consistently high UUAS scores} across all layers compared to the other two non-linear probes (Polynomial and Sigmoid). \cite{WhitePSC21} provides reasoning for this by showing a resemblance of RBF's structure with that of BERT's attention mechanism. BERT and BERT$_\text{LARGE}$'s UUAS score for the last layer is 69.71 and 60.77 respectively for RBF. The UUAS score's trend for a linear probe for both these models runs along with RBF's score with 66.65 and 60.10 for BERT and BERT$_\text{LARGE}$ respectively. Then the question arises, can we conclude that RBF Probe is better than Linear Probe? 

To investigate this further, we utilize the visualization technique we introduced in Section \ref{ssec:metric}. The $\text{strength}$ between two words (Equation \ref{eqn:8}) is shown by the gradient in Figure \ref{fig:sent1}. The edges predicted by probes having lesser strength are orange in color while the edges having greater strength have bluish color. We see in Figure \ref{fig:sent1} that linear probes predict more edge dependencies which are (1) \enquote{false negatives} (\textit{weekend --- due, to --- clogged}), (2) \enquote{true positives} but with a lower strength (\textit{I --- called, the --- weekend}). While the RBF probe comparatively predicts lesser edges but with higher strength and more correctness. We did the qualitative visualizations for multiple sentences at random and found similar patterns for them too. However, as future work, we can think of incorporating the strength with which two edges are connected in the calculation of UUAS scores. 

Furthermore, to investigate how dependency tree syntaxes are encoded in every layer, we applied all four probing variants to BERT and BERT$_{\text{LARGE}}$. However, since the RBF probe's performance is promising, we discuss the results of probing with this variant for BERT. Out of the 12 layers, we select three particular layers to discuss because of their functional position in the BERT model (lower, middle, and last layer) and also because of their high UUAS scores. The lower layers of the BERT try to capture linear word order \cite{Rogers2020API} and this is affirmed by our observation of Figure \ref{fig:a} as every other word is spanning out and connecting to every other word in its neighborhood. We see words from the first sub-part of the sentence (\textit{He listens and is excellent is diagnosing}), connecting with words from the second sub-part of the sentence (\textit{addressing and explaining}) which then is getting connected with the last sub-part of the sentence. 

Figure \ref{fig:b} shows the dependency tree of mid-layer 6 where many edges with high strength are predicted. The mid layers convey the most syntactic information as per \cite{Rogers2020API}. That might be true as we see stronger subject-verb agreement (\textit{he-listens}) but we also see wrong subject-verb pairs being predicted (\textit{diagnosing-exercises}). Layer 6 overall forms a lot of correct and incorrect dependencies and the high UUAS score can be attributed to the fact that due to so many connections being formed, some connections are predicted correctly but they might've been just formed \enquote{by chance} and do not have any meaning. \textbf{The UUAS score thus needs to also penalize such false-positive dependencies.}

Finally, we see that Figure \ref{fig:c} has very selective connections and the correct connections are predicted with high strength while the incorrect with low strength. This affirms with the observation in \cite{Rogers2020API} where the final layers are more task-specific. 

We also try to analyze whether the context is important for language models to form representations with syntax dependencies. We do this by running the same experiments but passing individual words of the sentence independently to obtain non-contextualized representations for them. We find that this degrades the performance and contexts are indeed important. Detailed discussion in Appendix \ref{appendix:ncr}.

\section{Conclusion}
In conclusion, we see that non-linear probes like RBF are better at describing the knowledge its embeddings have about linguistic properties like syntax trees. Our qualitative experiments make us realize that UUAS is not enough indicator of how good a probing design is, and incorporating the strength of predicted edges and false positives may make the UUAS more robust. Also, we see that layer-wise BERT gains knowledge functionally but our understanding of this is still in its infancy and more rigorous experiments are needed.  This we would like to address in our future work. 

\bibliography{acl_latex}

\clearpage

\appendix
\section{Non-contextualized representations} \label{appendix:ncr}

Figure \ref{fig:ncr} shows the performance of probes when word representations are taken independent of the sentence. From Section \ref{sec:discuss}, we see qualitatively that syntactic dependencies are more prominent in deeper layers. For BERT, the trend of Figure \ref{fig:bert} is decreasing for all probes. This shows that dependency structures between words are captured much better (from Figure \ref{fig:bbl}) within the context of a sentence.

For GPT2 though, results are more arbitrary. From Figure \ref{fig:gpt}, we observe that linear and polynomial probes have moderately high UUAS scores which are somewhat maintained across layers but UUAS scores from RBF and sigmoid probes are quite haphazard. Since GPT2 has been pre-trained on an autoregressive task of predicting the next token, it's only given the context of previous tokens which is a limited view than what BERT has. This might be the reason that the words have better non-contextualized representations. We couldn't provide intuition on why RBF and sigmoid probes don't work in this case and leave it for future work.

\begin{figure}[t]
    \centering   
    \includegraphics[width=0.5\textwidth]{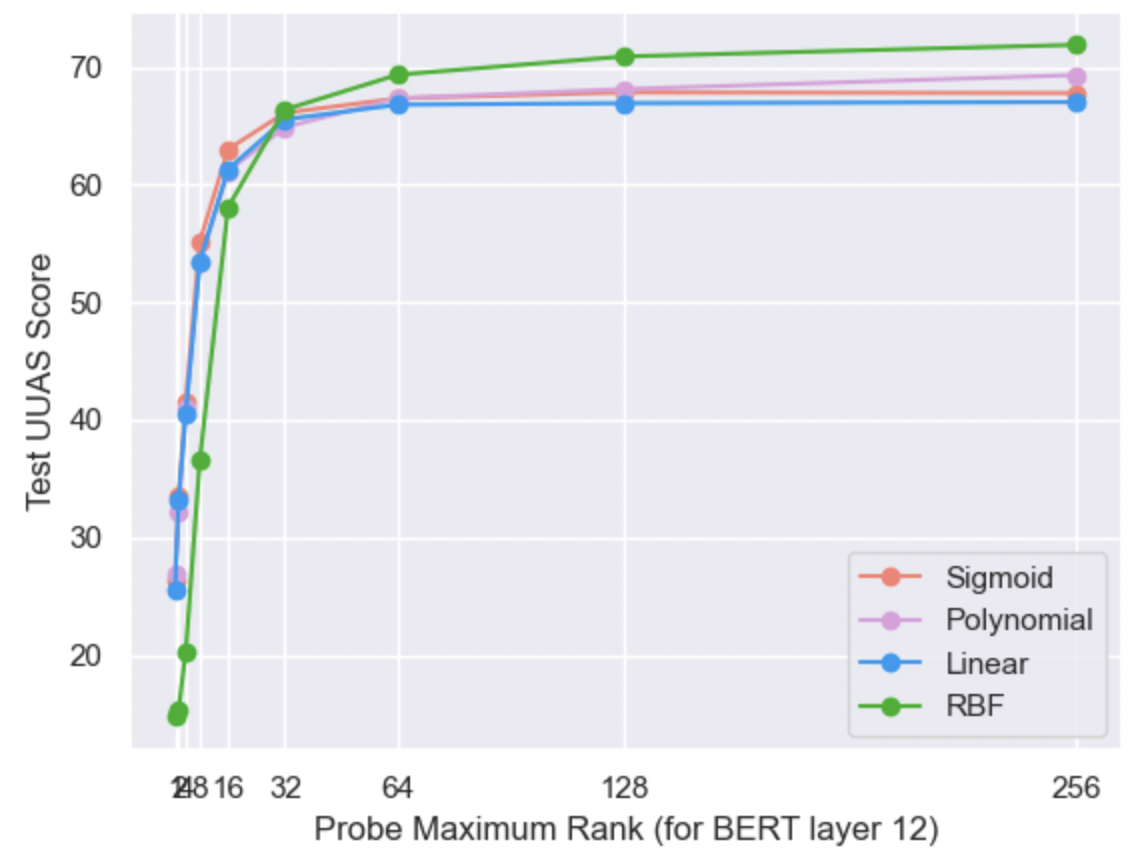}
    \caption{UUAS Scores for varying rank of matrix $B$ where rank $\in \{1, 2, 4, 8, 16, 32, 64, 128, 256 \}$}
    \label{fig:rank}
\end{figure}

\begin{figure*}[h]
  \centering
  \begin{subfigure}[l]{0.45\textwidth}
    \centering
    \includegraphics[width=\textwidth]{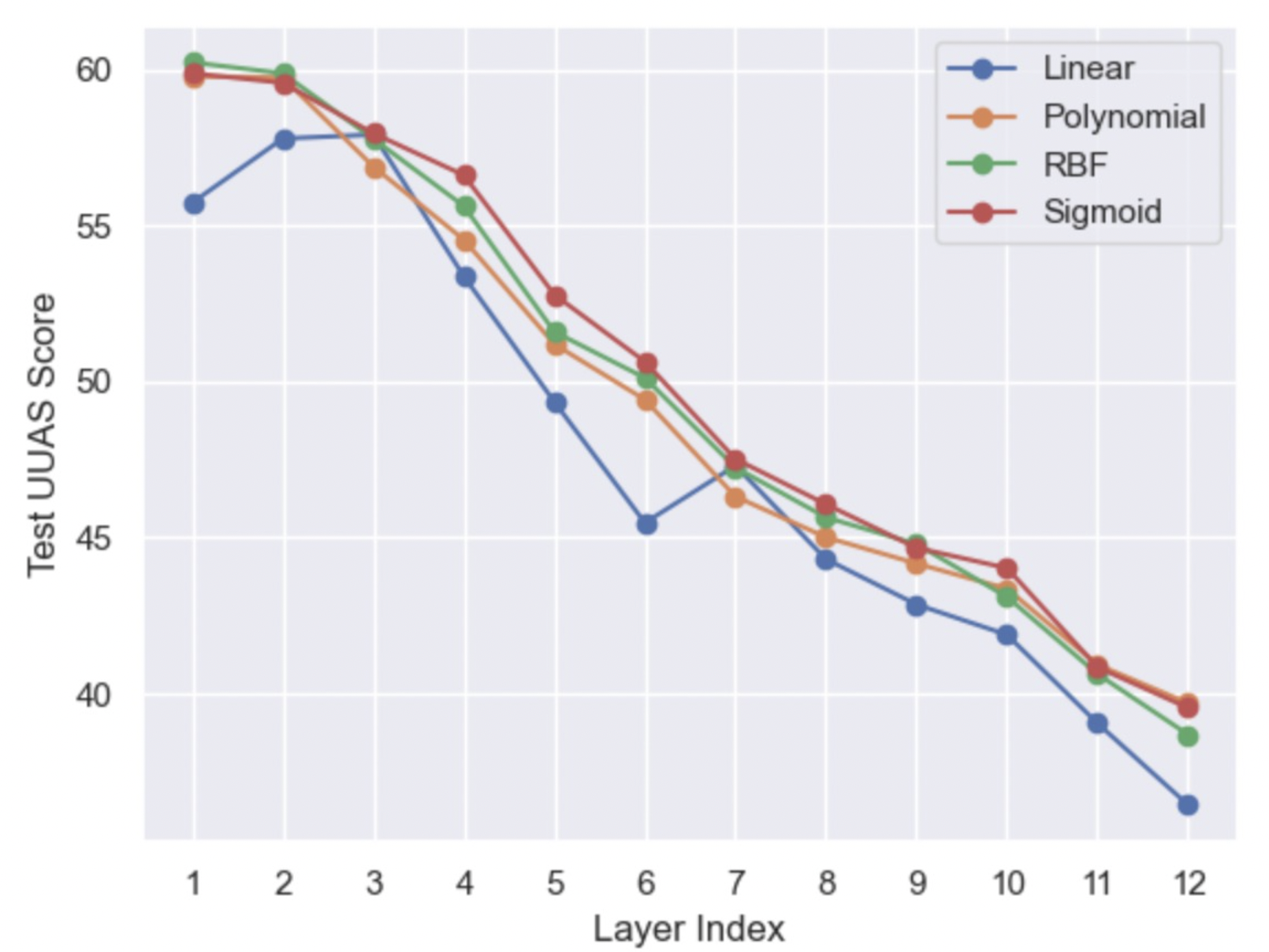}
    \caption{BERT} \label{fig:bert}
  \end{subfigure}
  ~
  \centering
  \begin{subfigure}[r]{0.45\textwidth}
    \centering
    \includegraphics[width=\textwidth]{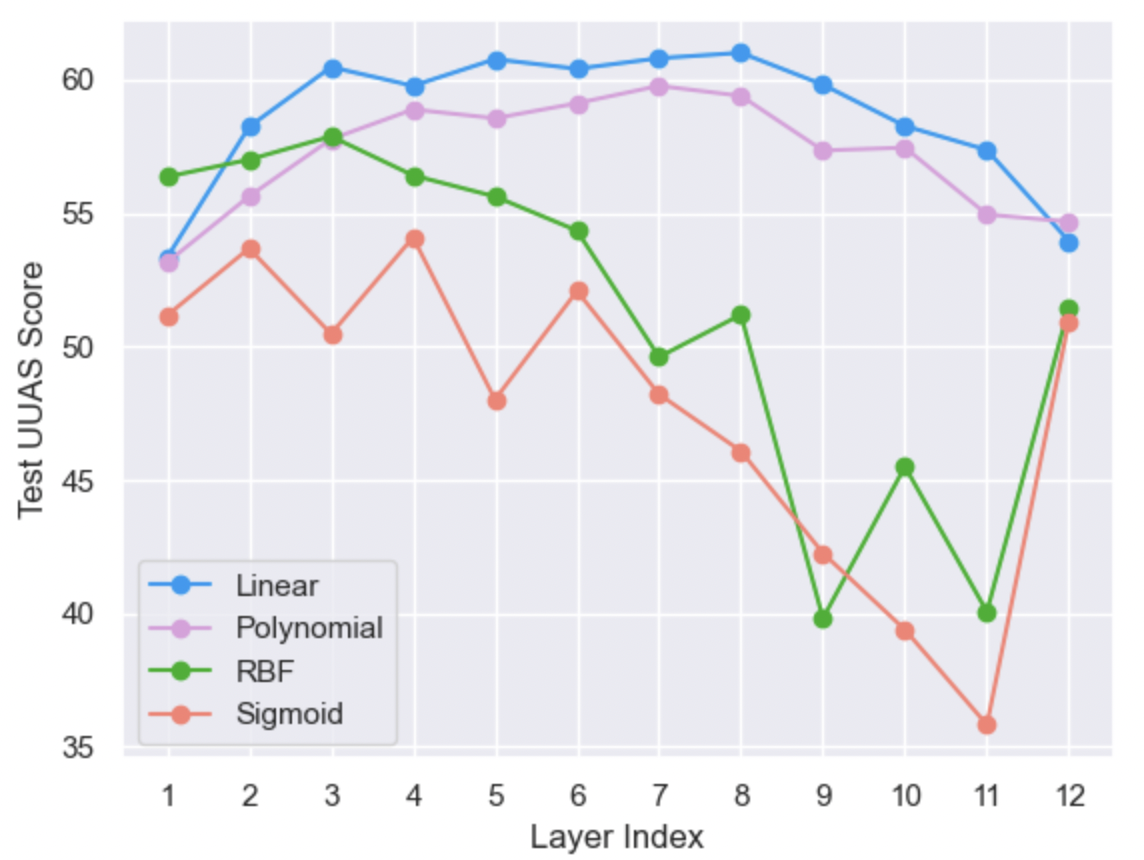}
    \caption{GPT2} \label{fig:gpt}
  \end{subfigure}
  \caption{UUAS Scores for non-contextualized representations} \label{fig:ncr}
\end{figure*}

\section{More Visualizations of Dependency Trees}
Figure \ref{fig:fig6} shows various dependency trees for BERT$_\text{LARGE}$. We do not observe gradual learning of syntax trees as we did for BERT for these particular layers. Due to time constraint, we couldn't generate visualizations for all of the 24 layers. And it might be the case that these particular layers we visualize might not depict the representative learnings. As part of future work, we  aim to investigate whether RBF exhibits gradual learning of syntax tree for BERT$_\text{LARGE}$ too. 

\section{Analysis of transformation rank}
We train our probes of varying $d$, that is, specifying a matrix $B \in \mathbb{R}^{d \times m}$. As shown in Figure \ref{fig:rank}, increasing $d$ beyond 64 or 128 leads to no further gains in UUAS.

\begin{figure*}
  \centering
  \begin{subfigure}[l]{0.45\textwidth}
    \centering
    \includegraphics[width=\textwidth]{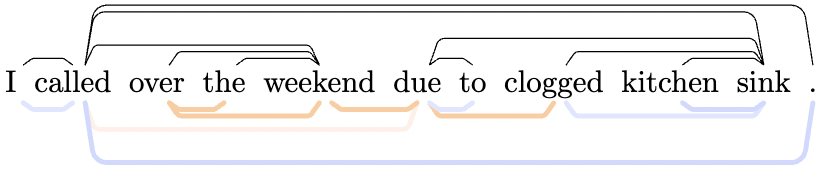}
    \caption{Polynomial}
  \end{subfigure}
  ~
  \centering
  \begin{subfigure}[r]{0.45\textwidth}
    \centering
    \includegraphics[width=\textwidth]{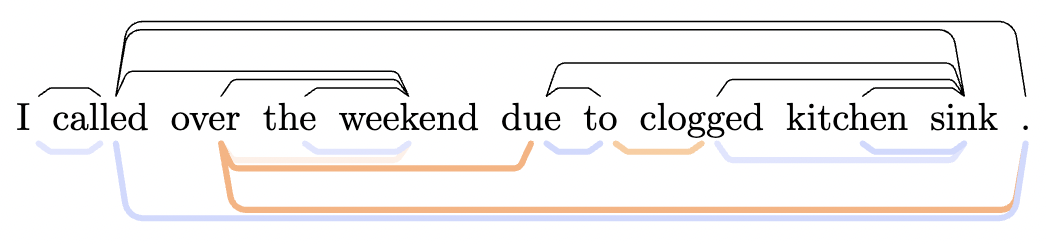}
    \caption{Sigmoid}
  \end{subfigure}
  \caption{Dependency trees for BERT Layer 12}
\end{figure*}

\begin{figure*}
  \centering
  \begin{subfigure}[l]{\textwidth}
    \centering
    \includegraphics[width=\textwidth]{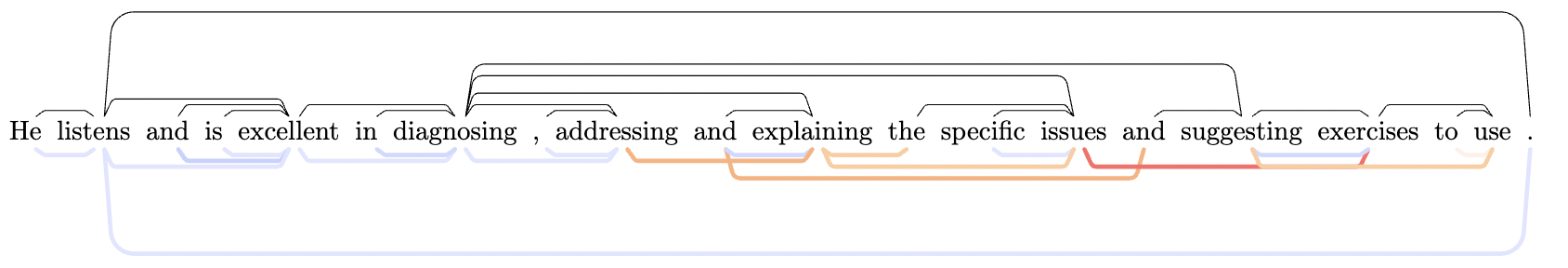}
    \caption{Layer 3}
  \end{subfigure}
  
  \centering
  \begin{subfigure}[l]{\textwidth}
    \centering
    \includegraphics[width=\textwidth]{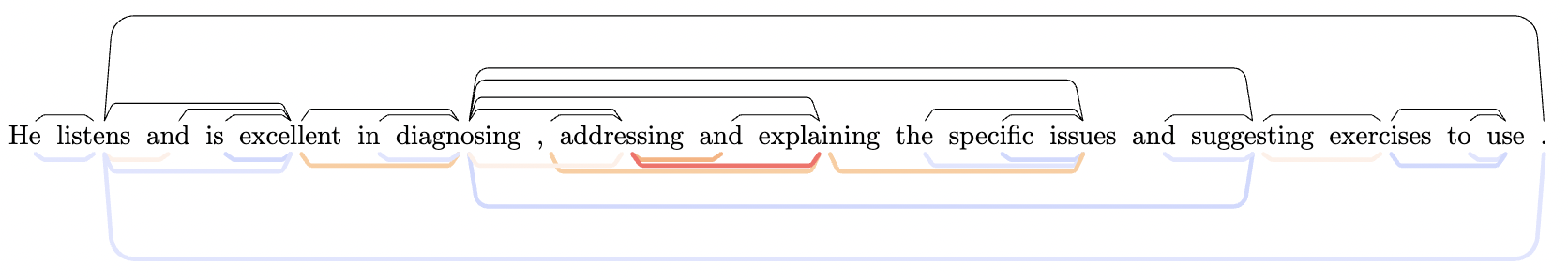}
    \caption{Layer 16}
  \end{subfigure}
  
  \centering
  \begin{subfigure}[l]{\textwidth}
    \centering
    \includegraphics[width=\textwidth]{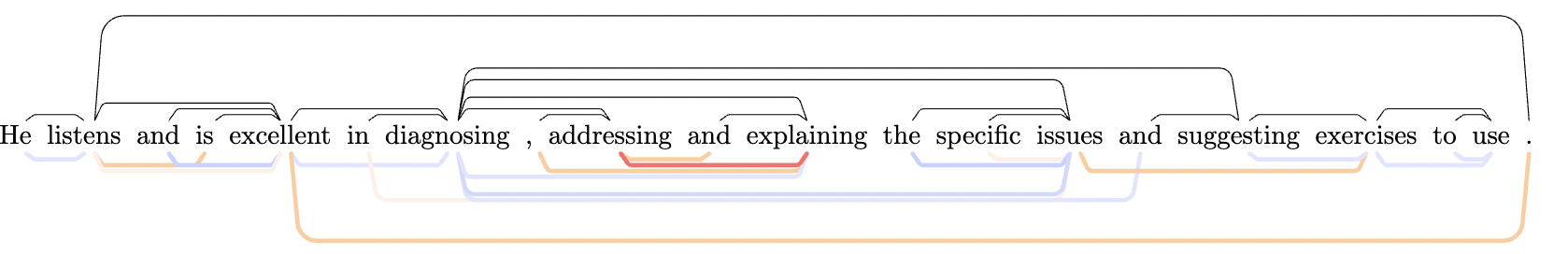}
    \caption{Layer 24}
  \end{subfigure}
  ~
  \caption{Dependency trees from predicted squared distances on BERT$_\text{LARGE}$ by RBF Probe for layers $\in \{3,16,24\}$} \label{fig:fig6}
\end{figure*}

\end{document}